# Human Activity Recognition on wrist-worn accelerometers using self-supervised neural networks

Authors: Niranjan Sridhar*, Lance Myers

Verily Life Sciences, LLC (Alphabet), South San Francisco, California, USA

**Address for correspondence:**
Niranjan Sridhar
Verily Life Sciences
269 E Grand Ave
South San Francisco, CA  94080-4804
nirsd@verily.com

# Abstract

Measures of Activity of Daily Living (ADL) are an important indicator of overall health but difficult to measure in-clinic. Automated and accurate human activity recognition (HAR) using wrist-worn accelerometers enables practical and cost efficient remote monitoring of ADL. Key obstacles in developing high quality HAR is the lack of large labeled datasets and the performance loss when applying models trained on small curated datasets to the continuous stream of heterogeneous data in real-life. In this work we design a self-supervised learning paradigm to create a robust representation of accelerometer data that can generalize across devices and subjects. We demonstrate that this representation can separate activities of daily living and achieve strong HAR accuracy (on multiple benchmark datasets) using very few labels. We also propose a segmentation algorithm which can identify segments of salient activity and boost HAR accuracy on continuous real-life data.

# Introduction

Assessments of activities of daily living (ADL) are used in the early detection, diagnosis, treatment, and care of many health conditions, including arthritis, depression, stroke, heart failure, Parkinson disease, dementia, and Alzheimer disease [1-6]. Further, in 2011, the United States National Health Interview Survey determined that 20.7% of adults aged ≥85 years, 7% of those aged 75-84 years, and 3.4% of those aged 65-74 years needed help with ADLs [7]. As the median ages of most nations continue to rise, accurate measurement of ADLs is critical for improving health outcomes and targeting effective interventions as well as decreasing the cost to patients and societies. Although numerous clinical diagnostic tests [8-10] include ADL assessments, these typically rely on self-reported data. While such data provide valuable perspectives, complementing them with objective measures would enable comparison of symptoms and outcomes over time or across different patients and allow the effects of interventions to be quantified.

A prerequisite to developing novel objective measures of functional capacity during ADL is the detection and recognition of those activities in real-life settings. Recent advances in wearable technology have enabled the deployment of algorithms capable of measuring heart rate, monitoring individual activity levels, counting steps, and tracking sleep from wrist-worn devices [11]. Such measures allow consumers to track their health metrics over time and are emerging as valuable digital biomarkers in clinical studies [12]. A key sensor found in most wrist-worn wearables is the 3-axis accelerometer, or inertial measurement units (IMU) sensor, which can enable a wide variety of health-related assessments about individual users, including automated, sensor-based human activity recognition (HAR).

Attempts to develop HAR models on smartwatches, smartphones, other body-worn sensors, or via radar, video, and other ambient sensors, have been hampered by a lack of large datasets - ones with high-quality reference activity labels in real-life settings - that can be used to build and test models. Collection of HAR data in real-world settings requires attaching sensors to study participants inside their home or workplace, which can impose significant burdens. In addition, labeling HAR data usually requires participants to self-report their activities, introducing variability and noise due to participant subjectivity and compliance. Independent reviewer annotation can be done if the study includes video surveillance data, however this raises additional privacy concerns.

Due to these obstacles, most HAR benchmark datasets [13-16] contain data on 1-20 research participants performing 1-15 activities for 1-5 minutes under controlled or clinic-like conditions. Given the small size and contrived nature of these datasets, models trained and tested on them are unlikely to generalize to new participants, devices, and activities, especially in uncontrolled settings. While many published studies report near 100% accuracy on these datasets individually, cross-dataset performance has not approached such levels of accuracy, prompting debate about whether accelerometer-based HAR is even possible [17]. Therefore, novel methods for building robust and practical HAR models that do not require extensive labeling of large datasets are needed.

The Project Baseline Health Study (PBHS) [18] is a prospective, multi-center, longitudinal study that aims to map human health through a comprehensive understanding of the health of an individual and how it relates to the broader population. As part of the PBHS, participants agree to wear a wrist-worn device during daily activities that continuously collects high-resolution IMU data. In our study, we used these data to create a representation of activity without any labels. We demonstrated the efficacy and generalizability of this feature representation on HAR using 4 publicly available benchmark ADL datasets [13-16]. We further designed a segmentation algorithm to identify periods of well-defined activities in a continuous data stream of free-living data and demonstrated its efficacy on the PAMAP2 dataset [13.

# Results

The primary result of this study is our activity representation model, which we evaluated for label-efficient activity recognition accuracy using public benchmark datasets. Our secondary result is our activity segmentation model, which acts on the representation model outputs. This model can be evaluated on a dataset containing continuous streams of IMU data.

## Unsupervised representation learning setup

We trained a convolutional neural network encoder on wrist-worn accelerometer data gathered from the PBHS study. The encoder 'encodes' a given 10-second window of 3-axis accelerometer data into a 256-dimensional feature vector.

During the training process, we used a contrastive loss objective to train the encoder. The contrastive objective is a binary classification objective that forces the model to distinguish between 2 types of pairs of inputs. We created 'positive' pairs of 10-second windows that are related to each other because they belong to the same subject and are either neighboring windows or augmented transforms of each other (See Methods for details). 'Negative' pairs are unrelated windows, either distant in time or belonging to different subjects altogether. These labels, which are created with no external domain-specific information and thus 'self-supervised,' cause the encoder to extract features that are invariant to noise and can characterize motion that is consistent over time.

During inference, we drop the contrastive loss network, and the encoder takes as input a single 10-second window and returns a 256-element feature vector or embedding. Figure 1 shows the architecture of the full training pipeline, specifically the encoder model. Supplementary Figure 1 shows t-SNE representations of the embeddings of all 10-second windows from each benchmark dataset. We see qualitatively that different activities are separated into clusters in the embedding space.

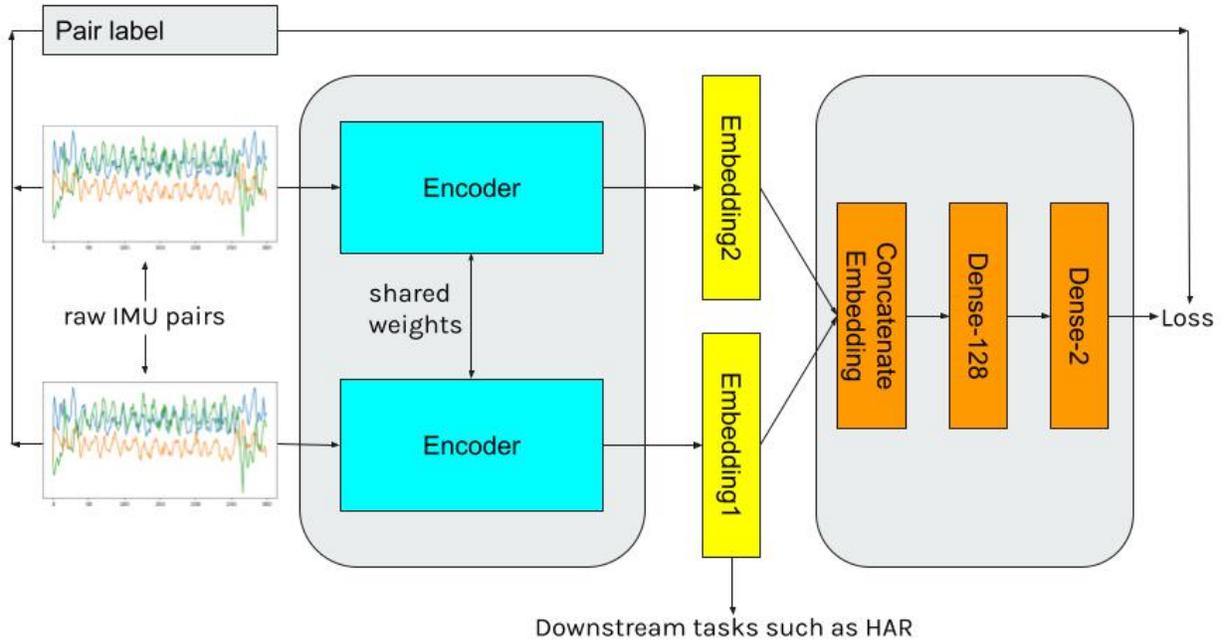

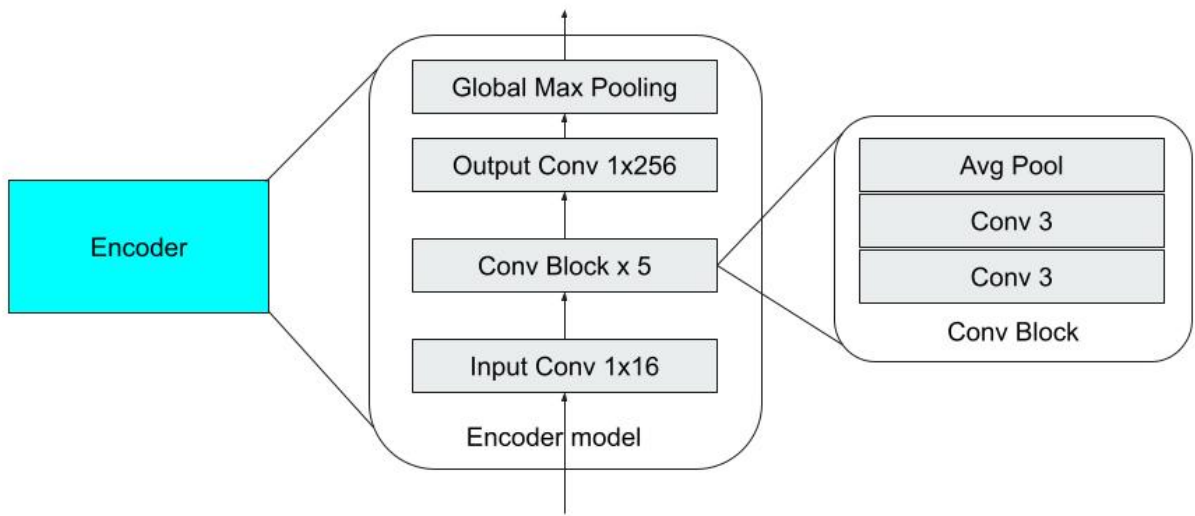

Fig 1. Model training pipeline. The main model is the convolutional neural network encoder which takes 10 second accelerometer windows as input and returns a vector embedding. The auxiliary dense layers are added during training to predict which pairs of IMU windows are labeled as positive.

# Activity recognition on benchmark datasets

## *Label-efficient activity recognition*

To quantify the descriptive and generalization properties of our representation, we evaluated it on the downstream task of activity classification on 4 publicly available benchmark datasets. To do this, we first encoded the accelerometer data in the datasets into embedding vectors by splitting them into non-overlapping 10-second windows and applying the pretrained encoder on them. We then split each dataset into a random 75:25 train-and-test split. Finally, we trained logistic regression models on subsets of the training set where we randomly sample $n \epsilon$ {1, 5, 10, 15, 25, 50} windows from each class with replacement from the training set and evaluated these models on the test set. This process was repeated 10 times to derive confidence intervals.

Figure 2 shows the HAR accuracy of 4 datasets as a function of *n*, the number of labeled examples per class. In the low-label regime of n<20, we see large improvements in classification accuracy with each additional labeled example per class in all datasets. We also added a comparative baseline showing the classification performance achieved by replacing the embedding with an 8-dimensional feature vector containing the means and standard deviations of the 3 axes and norm of the IMU. (Baseline model here and elsewhere in this manuscript refers to the HAR model trained using simple statistical features instead of our learned representation. Project Baseline Health Study will always be referred to in full or as PBHS.)

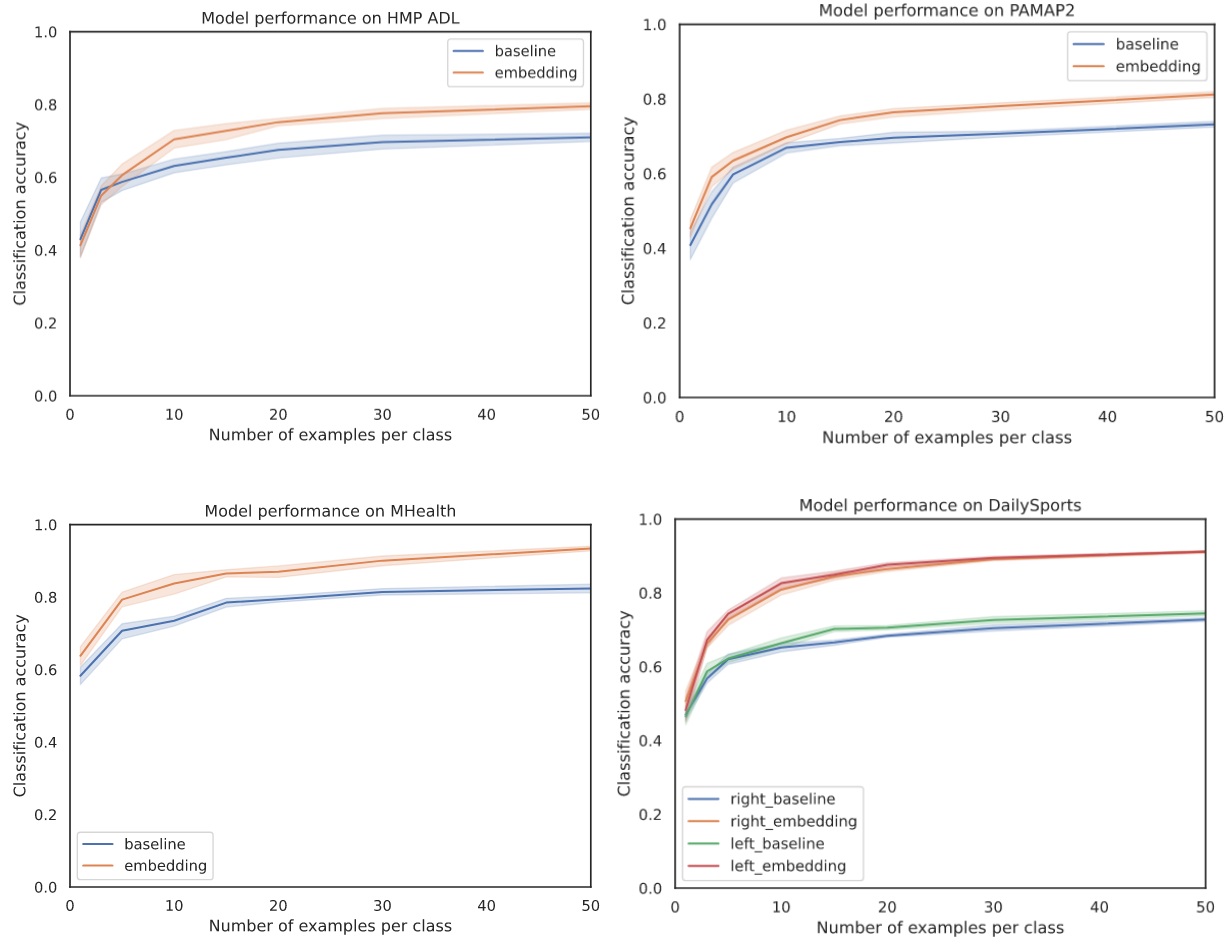

Fig 2. Label efficiency of window-level classification accuracy on benchmark datasets.

# Activity recognition in real-life settings

### *Salient activity segmentation*

We apply an unsupervised segmentation algorithm on the embedding vector time series to isolate periods of distinct and consistent activity from a continuous stream of unlabeled data. The algorithm identifies sequences of windows whose representations are similar to each other and dissimilar to windows outside the sequence (See Methods for more details). We tested this on the PAMAP2 dataset but this time included the null class, which comprises amorphous activities that may or may not fall in any class. With this inclusion, each participant represented in this dataset now had a continuous stream of activity data with labeled activities some of which are separated by unlabeled or 'null' motion. The segmentation algorithm takes a timeseries of embedding vectors and returns a list of paired start and end times. Each start and end time pair identifies a segment predicted to be one contiguous activity. This prediction is completely unsupervised and agnostic to the type of activity. Figure 3 shows all the labeled activities and predicted salient activity segments of one subject in the PAMAP2 dataset.

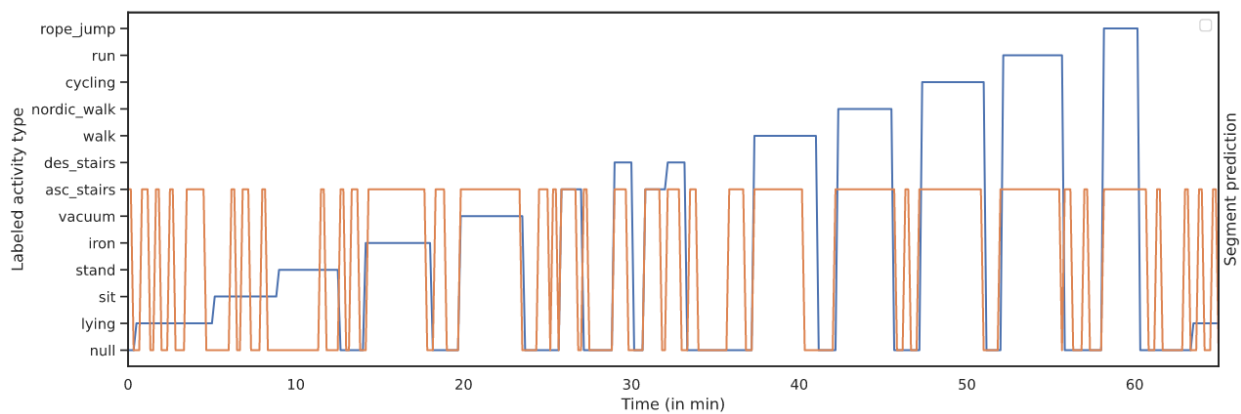

Fig 3. Salient activity segmentation. The blue curve indicates the labeled activity type noted on the left y-axis. The 'null' class is not a well-defined activity and fills time between designated activities such as walking and lying. The non-zero orange segments indicate segments we predict as being a continuous activity, irrespective of the type of activity.

We measure the performance of our segmentation algorithm by using 2 sets of measures: event-level precision and recall capture the fraction of the total number of predicted segments that contain only 1 labeled activity, and the fraction of the total number of labeled activity segments that overlap with at least 1 predicted segment, respectively. Window-level precision and recall capture the fraction of the total duration of predicted segments that contain only 1 labeled activity and the fraction of the total duration of labeled segments that overlap with predicted segments. We observed a recall/precision of 1.0/0.99 on the event level and a recall/precision of 0.66/1.0 at the window level.
Performance metrics separated by activity labels are summarized in Table 1.

| Activity | Event | | Window | |
|---|---|---|---|---|
| | Precision | Recall | Precision | Recall |
| Lying | 1. | 1. | 1. | 0.61 |
| Sit | 1. | 1. | 1. | 0.34 |
| Stand | 1. | 1. | 1. | 0.37 |
| Iron | 1. | 1. | 1. | 0.38 |
| Vacuum | 1. | 1. | 1. | 0.90 |
| Ascend stairs | 1. | 1. | 1. | 0.74 |
| Descend stairs | 0.89 | 1. | 0.96 | 0.65 |
| Walk | 1. | 1. | 1. | 0.92 |
| Nordic walk | 1. | 1. | 1. | 0.99 |
| Cycling | 1. | 1. | 1. | 0.96 |
| Run | 1. | 1. | 1. | 0.94 |
| Rope jump | 1. | 1. | 1. | 0.73 |

Table 1. Precision and Recall of the salient activity segmentation algorithm at the event and window level. Event-level precision and recall capture the fraction of the total number of predicted segments that contain only 1 true activity, and the fraction of the total number of labeled activity segments that overlap with at least 1 predicted segment, respectively. Window-level precision and recall capture the fraction of the total duration of predicted segments that contain only 1 labeled activity and the fraction of the total duration of labeled segments that overlap with predicted segments.

*Salient activity recognition*

Next, we applied the window-level activity classification analysis on PAMAP2 again, this time after filtering out non-salient activities and keeping only windows that belonged to the segments described above. We found that this improved performance substantially, especially for the representation model with very few labels (n<10), see Figure 4.

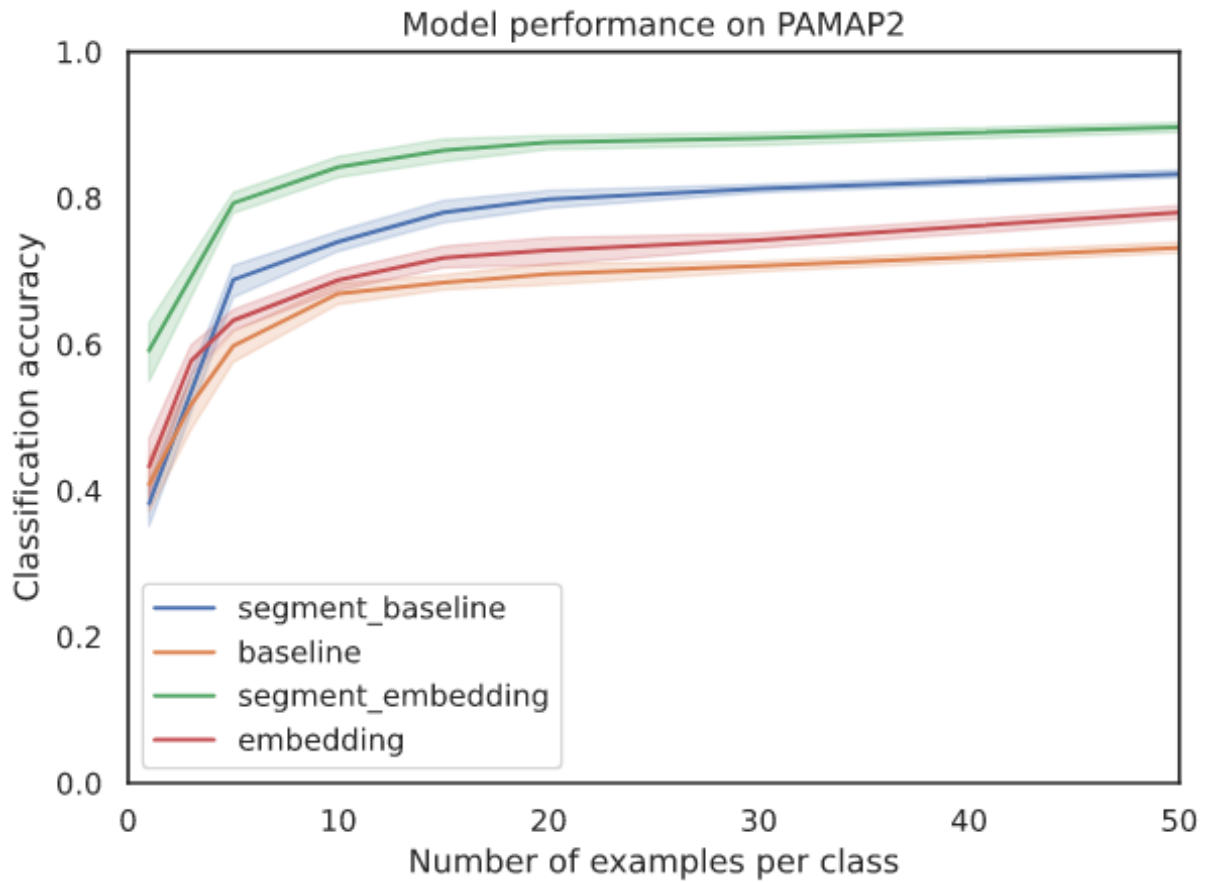

Fig 4. HAR accuracy on PAMAP2 dataset. The 'segment' prefixed curves are measured by filtering out windows that do not belong to the predicted salient activity segments. Filtering improves the baseline model accuracy, but improves the embedding model more especially for fewer than 10 labels.

Without the salient activity filtering on the PAMAP2 dataset, the HAR performance of the embedding for n<10 labeled examples per class showed only modest improvement over the trivial features baseline. In contrast, on the other 3 benchmark datasets, the embedding features show large improvement over the trivial baseline. When we filter out the salient activities, HAR performance for the embedding model improves greatly especially in the low-label regime.

## Discussion

*Comparison with state of the art*

Most published works we reviewed performed training and testing of the entire network on every dataset individually, often achieving very high accuracy. We expect that such models are unlikely to generalize, hence we do not compare our work to efforts that only ran their train-and-test on the same dataset and limit our comparisons to published work where 2 or more HAR datasets were tested on the same model or representation. Self-supervised models are a practical way to make highly robust and generalizable models that do not overfit on small datasets when pre-trained on large unlabeled datasets [23].

In addition, some published works pre-train a network without labels and evaluate on multiple benchmark datasets, however they either fine-tune the entire network on the labeled datasets or use a multilayer network on top of the embeddings as the classification model. Given the high predictive power of even trivial features as seen in our baseline model, a complex model can easily produce very high accuracies on any of these datasets, yet not be generalizable to multiple datasets. In our study, we use a simple logistic regression model as our prediction head because a linear model cannot create new feature combinations and can test the predictive power of features learnt by our pre-trained activity representation.

All benchmark datasets we show here contain multiple body-worn sensors at locations including arm, chest, lower back, and thigh. Many published works train and test models on multiple combinations of sensors, and often find that HAR accuracy using wrist-worn sensors is much lower than that achieved using multiple body-worn sensors. We focused our efforts on wrist-worn sensors because they represent a mature consumer health technology already in daily use by hundreds of millions of people, thus supporting the near-term practical utility of our approach.

Many of the comparable published reports describe using overlapping windows for both training and evaluation. While this increases the size of the dataset for training, it also makes the windows highly correlated with each other, and its use in evaluations could result in spuriously high numbers. To avoid this, we use a conservative approach and only use non-overlapping windows in our evaluations.

Many previous works also did not compare their learned representation against good baselines to quantify the predictive power of their features over simpler methods. Our 'trivial' feature vector uses simple statistical measures of the input signal and produces a strong baseline on all benchmark datasets, outperforming even some of the published results.

In summary, in addition to our neural network and classification models, we have proposed and implemented a number of methodological improvements to the process of benchmarking HAR performance. We hope that the HAR research community can benefit from these recommendations and progress towards publishing reliable benchmarks that can be easily compared leading to faster and more reproducible research.

| Authors (year) | HMP | PAMAP2 | MHealth | DailySports | Other |
|---|---|---|---|---|---|
| J. Wang et. al (2018) [19] | - | 39.21 | - | 57.97 | |
| Xin Qin et al (2020)** [22] | | 63.9 | | 60.7 | |
| Holzemann et al (2020) [21] | | 20.1 | | | 42.8 |
| TransAct (2017) ** [20] | | | 82 | 85 | |

| | | | | |
|---|---|---|---|---|
| Ours (baseline) | 70.9 | 74.3 | 82.4 | 72.8 |
| Ours (embedding) | 79.5 | 83.3 | 93.4 | 91.1 |

Table 5. Lists random test set activity classification accuracies achieved by comparable efforts that show generalization of models across datasets. ** indicates the performance is measured on a smaller set of activities than what is found in the dataset.

## Practical utility of unsupervised segmentation

In many comparable published works, the label efficiency of learnt representations is evaluated by sampling labels from a benchmark dataset as we have also done in the Results section. However, this is not a realistic simulation of a practical dataset of sparse labels. Although we constrained the number of labels, the quality of the labels is always perfect; as we increased the number of sampled labels, we started sampling multiple high-quality labels from the same activity event. Such a scenario can only occur in controlled environments and is unlikely in real-life data collection. We present a more practical scenario in which participants wear the watch in daily life and occasionally tag the activity they are engaged in. In this scenario, if participants tag only once while doing a certain activity, the result is only 1 label per event. If instead participants tag the start and end times of activities as in our study, we obtain more labels but at the cost of substantial noise, as subjects often forget to end the tag correctly or have unplanned interruptions and resumptions of their activities that are not captured in the tags. As a result, the number and quality of labels is much lower in practical data collection than can be simulated by sampling.

Our unsupervised segmentation algorithm can identify contiguous segments of activity in a real-life datastream without using any labels. As we show on the PAMAP2 dataset, this algorithm has very high precision: thus if we have a single high-quality label within a segment, we can expand the label to the whole segment thus increasing the efficiency of a very small number of labels. Alternatively, the algorithm can also be used to isolate the dominant activity segment and reject neighboring periods that are dissimilar to the dominant activity (Supplementary Figure 2). This can be used to improve the detection of specific activities amid noisy activities and drop incorrect labels in a free-living dataset.

*Limitations*

There are a few limitations of our proposed HAR scheme. First, half of the training objective of our self-supervised loss assumes that neighboring windows are likely to be part of the same activity. This objective enables the encoder to learn a rich representation for 'slow-moving' activities (i.e. activities that have the same underlying structure for at least 1-5 minutes). However, this objective could potentially act against learning rich representations of activities that last less than 1 window (e.g. standing up from a chair or opening doors). The second training objective, which uses augmentations instead of temporal proximity for defining coincidence pairs, is better suited to represent these activities. More analysis is needed to quantify the impact of the 2 objectives and how they affect different activities. This will require more detailed studies with many more activities than are found in the benchmark datasets.

The unsupervised segmentation algorithm isolates periods of stable salient activity in an continuous IMU datastream. Using this as a filter to improve label quality can potentially remove noisy but correctly labeled data. For example, in Supplementary Figure 2 we see a 15 minute period labeled by a participant as 'exercise'. The segmentation algorithm identifies multiple segments of activity which stand out distinctly from the intervening motion. The windows in the time between two segments are not as similar to each other as the windows within the segments. However, it is likely that the intervening periods are still correctly labeled as 'exercise' even though they may be a less contiguous form of exercise than the segments. Therefore if we filter windows using salience segmentation to improve label quality, we can get clean examples of activities with high precision, at the expense of lower recall.

At a broader level, the clinical applications of HAR are not yet fully understood. As mentioned earlier, numerous clinical scores use self-reported surveys as measures of functional capacity. Remote sensor-based HAR could potentially be used to develop objective measures of functional capacity; however, testing this hypothesis requires robust HAR in a real-life scenario, which has not been demonstrated yet. We hope that our work here can enable studies that can deploy remote HAR in patient populations to measure and improve patient outcomes.

Finally, a pervasive critique of deep learning systems concerns the lack of explainability of features learned by these networks. In a sense, self-supervised models are a useful tool for investigating explainability as they are often more robust than fully supervised learning [23]. However this is still an area of active research and beyond the scope of this work.

# Methods

## Datasets

### *Self-supervised training*

The training dataset was a 1-month period of free-living data collected in the PBHS [18] via a smartwatch (Verily Study Watch) equipped with an accelerometer. No filtering or inclusion/exclusion criteria were applied, yielding 42,000 hours (~15 million windows) of accelerometer data from approximately 1200 study participants. During model training, no labels of activity or health were used and the only context used to identify accelerometer data was the device identifier and timestamp.

## Testing (benchmark HAR)

We used 4 public datasets incorporating labeled activities of daily living: the HMPADL, PAMAP2, MHealth, and Daily Sports datasets [13-16]. The datasets are available at the UCI Machine Learning Repository http://archive.ics.uci.edu/ml/index.php. Each dataset contains the wrist-worn accelerometer data of a number of subjects who perform a predetermined set of activities. They also contain the start and end time of each activity. Some of the datasets also contain other data types such as accelerometers on the wrist or back which we do not use in this work. Table 2 summarizes the metadata of the benchmark datasets. Table 3 lists the activity types and the number of windows of each activity in the benchmark datasets.

| DATASET | Number of activities | Number of subjects | Device type | Sampling rate (Hz) | Wrist side |
|---|---|---|---|---|---|
| MHealth | 12 | 10 | Shimmer2 | 50 | right |
| PAMAP2 | 18 | 9 | Colibri | 100 | dominant |
| HMPADL | 14 | 16 | Unknown | 32 | right |
| DailySports | 19 | 8 | Xsens MTx | 25 | left, right |

Table 2. Summary of benchmark datasets metadata.

| Activity | Number of windows | | | |
|---|---|---|---|---|
| | MHealth | PAMAP2 | HMPADL | DailySports |
| Null | | 923 | | |

| Activity | | | | |
|---|---|---|---|---|
| Stairs (ascending + descending) | 60 | 115+106 | 83+32 | 232+232 |
| Ironing | | 240 | | |
| Vacuum | | 175 | | |
| Lying | | 195 | | 464 |
| Sit | 61 | 185 | | 232 |
| Stand | 60 | 188 | | 232 |
| Walk | 61 | 241 | 231 | 232 |
| Nordic Walk | | 188 | | |
| Cycling | 62 | 164 | | 464 |
| Run | 61 | 100 | | 232 |
| Rope Jump | | 48 | | |
| Jump | 20 | | | 232 |
| Arms up | 58 | | | |
| Crouch | 59 | | | |
| Jogging | 61 | | | |
| Waist bend | 58 | | | |
| Elevator | | | | 232+232 |
| (move+still) | | | | |
| Treadmill (flat + incline) | | | | 232+232 |
| Stepper | | | | 232 |
| Crosstrainer | | | | 232 |
| Basketball | | | | 232 |
| Rowing | | | | 232 |
| Brush teeth | | | 85 | |
| Comb hair | | | 58 | |
| Drink | | | 88 | |
| Eat | | | 115 | |
| Get up (+Lie down) | | | 96+23 | |
| Phone | | | 41 | |
| Sit down (+ stand up) | | | 23 + 22 | |
| Pour water | | | 117 | |

Table 3. Distribution of activities in benchmark datasets.

# Activity representation

## *Data preprocessing*

Because the datasets we used have data from different devices and different sampling rates, we first standardized input data by normalizing all values to units of g and resampling all time series to 30Hz. Resampling with and without anti-aliasing did not result in any notable difference in performance. A sampling rate of 30 Hz was chosen because 1) the training dataset (PBHS) is natively sampled at 30 Hz, 2) a literature survey shows that most activities of daily living are captured under a frequency limit of 15 Hz, and 3) smaller sampling rate implies smaller datasets and faster training/inference.

Next, we divided the time series into non-overlapping windows of 10 seconds each. Each window contains 3 time series corresponding to "acceleration_x", "acceleration_y" and "acceleration_z" together with the device identifier and the starting timestamp of the window.

## *Coincident pair sampling*

Next, we created coincident pairs. The underlying principle of coincident learning is to learn a mapping from input (in this case, accelerometer windows) to an embedding vector space in which coincident pairs of windows are closer to each other than pairs that are not coincident. Thus the choice of coincidence criteria decides the geometrical relationship between points in the embedding space. In this work, 2 main forms of coincidence pairs were used:

a) Temporal proximity: two windows were considered as coincident if they belong to the same user and occur within a specified time $\Delta t$. The temporal proximity criterion creates clusters of 'slow-moving' activities (i.e. activities that have the same underlying structure over periods as long as or longer than $\Delta t$). For this study, we chose a temporal proximity distance $\Delta t$ = 60 seconds.

b) Augmentation: a transformation was applied on the accelerometer signal and the original and augmented views were considered coincident. This criterion makes the representation invariant to certain perturbations that constitute common forms of noise, or variations seen within and between wrist-worn accelerometers. For example, we added random jump in baselines, baseline wander, median filtering, rotation in the x-y plane and Gaussian noise. For each of these augmentations and their compositions, we made pairs using the original window and its augmented version. Supplementary Figure 3 shows the transformations we used, along with examples of their effect on the data.

## Model Training

We used coincident pairs to train an encoder. The encoder architecture of the model is a convolutional tower with *c* convolutional blocks, followed by a dense layer with *d*-dimensional embedding output. We experimented with various hyperparameters and achieved stable results with *c=5* convolutional blocks and the *d=256* embedding size.

The construction of the optimization objective was inspired by the recent work in images and audio [24-25]. During training, the pairs were stacked in batch dimension and fed to the encoder to get embeddings for all windows (i.e., if the original batch size is *b=128*, effective batch size to encoder is *2\*b = 256*). We then created paired embeddings by concatenating the embeddings of all possible pairs of windows in the batch (i.e. *(2\*b)\*(2\*b) = 65,536* paired embeddings).

The resulting paired embeddings were then fed into a 2-layer perceptron projector. The output of the perceptron is a binary softmax prediction, with negative or class 0 indicating the two windows are not coincident and positive or class 1 indicating that the two windows are coincident. Figure 6 depicts how positive and negative pairs were assigned during training. There are *b* (*2 permutations) pairs with positive labels; these were the coincident pairs created in the previous step. There are also *2\*b* pairs that are identical (pairs of the window with itself).

The remaining *4*b*b* - *4*b* pairs are random combinations of windows sampled from the dataset and therefore labeled negative. Thus, while we explicitly created positive pairs, pairs created by two randomly picked windows from the dataset are labeled negative. This is a valid approximation that holds for large datasets where the number of subjects > *b* and the number of windows per subject >>*b*. Since there were many more negative pairs than positive, we down-weighted negative labels by a factor of *(2*b-2)* and downweighted the identity pairs to 0. The training loss was a cross-entropy loss between the predicted binary prediction and the coincidence labels. Training was continued for 0.5 million steps and the model was saved.

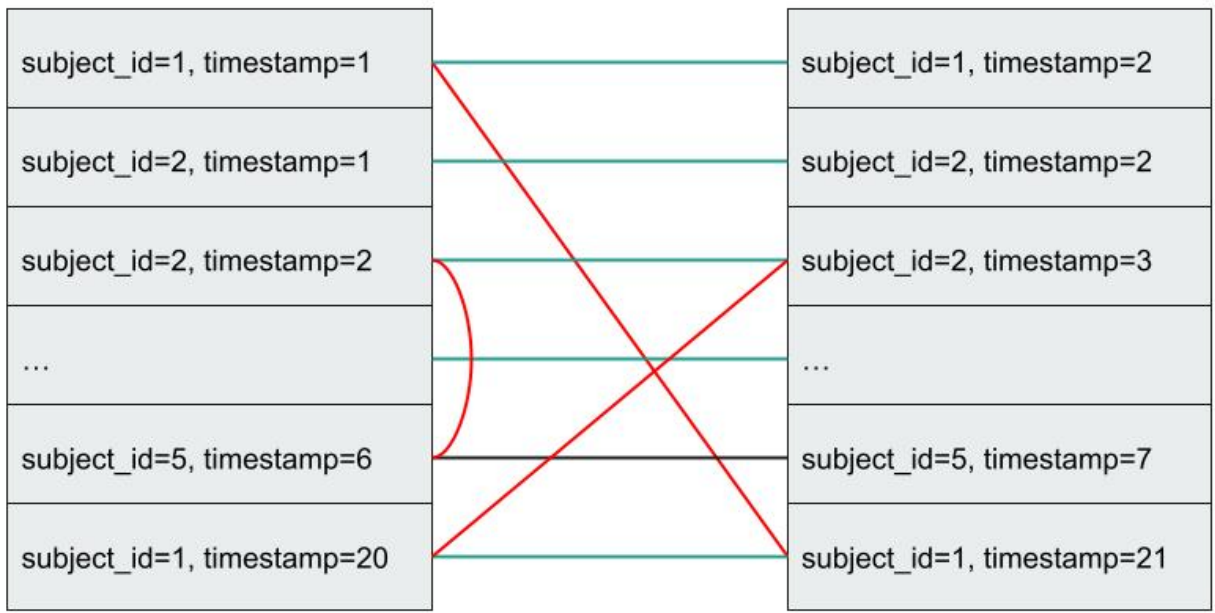

Fig. 6 Training inputs are batches of paired windows. Pairs *ii* (connected by green lines) for *i*th element of the batch are positive pairs that were sampled before training. Pairs *ij* (connected by red lines) for *i != j* are random combinations. For a large shuffled dataset, these combinations can be assumed to be negative pairs.

## Label-efficient HAR

We evaluated the quality of our encoder feature representation in 4 independent public datasets on the task of activity classification with few labels. In some datasets, there was a null class that did not correspond to any one activity and was filtered out for this analysis. During inference, we only used the encoder part of the model to infer over all the data, yielding one 256-element embedding vector for every 10-second window. We then divided each embedding dataset into 'train' and 'test' and randomly sampled *n* labeled windows for each class from the train dataset where n ϵ {1, 5, 10, 15, 25, 50}. Finally for each sampled dataset, we trained a logistic regression to predict the activity label and evaluate the model performance on the test set. We repeated this random sampling, train, and evaluation process 10 times for each set of parameters to obtain confidence intervals of model performance.

## Salient Activity Segmentation

For unsupervised segmentation of the embedding time series, we employed a region proposal network (RPN) with non-maximal suppression, inspired by the well-known Faster R-CNN object detection work [26]. However, because we did not have human-annotated truth labels to identify the correct regions, we disconnected the RPN segmentation from the unsupervised representation learning and instead we imposed a simple heuristic for selecting segments: the difference between the mean cosine similarity inside the segment and its neighborhood must be greater than two standard deviations of the cosine similarity inside the segment.
The salience of a segment *seg* with a neighborhood *nbd* is given by,

$$salience = Mean\ (sim(seg+nbd) - sim(seg)) - 2 * Stdev\ (sim(seg))$$

where *sim(seg)* is the set of all cosine similarities of all pairs of windows within the segment *seg*, *Mean (x)* is the mean of *x*, and *Stdev(x)* is the standard deviation of *x*.

To implement this, first we divided the continuous time series into 30-minute blocks or 180 windows (this was done to accommodate CPU memory constraints and had minimal effect on model performance). We have 2 natural constraints on our region proposals : a) temporal segments extend only in 1 dimension (time) and b) activities cannot occlude other activities;

therefore, each window can belong only to one segment. Thus, the RPN starts by creating all possible proposals of segments that can only have 180*179/2=16,110 unique values. First we apply a minimum length for the segments and discard all segments smaller than 30 seconds. For each proposal we define its neighborhood, which is the period before and after the proposal segment, each of length equal to half the segment's length or 5 minutes (30 windows), whichever is smaller. We can then measure the salience of the segment as defined above. Any segment proposals with a salience greater than zero are accepted as valid activity segments and the rest are discarded. We then applied a post-processing step similar to non-maximal suppression to combine any proposals that overlapped >50% and truncated any regions that overlapped <50% to remove all overlaps between segments. Finally we combined the 30-minute blocks back to obtain the set of discrete activity segments over the entire time series.

# Supplementary Figures

1.

HMPADL

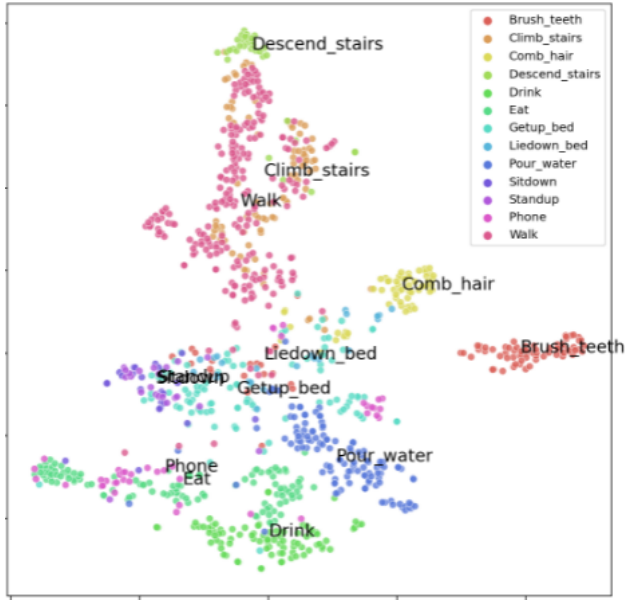

PAMAP2

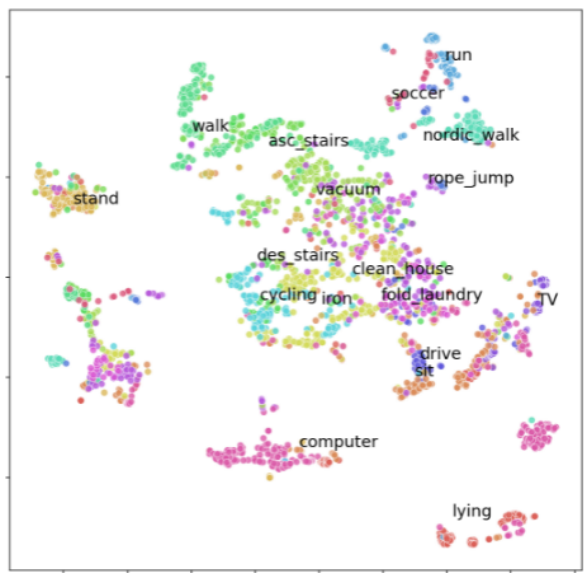

MHealth

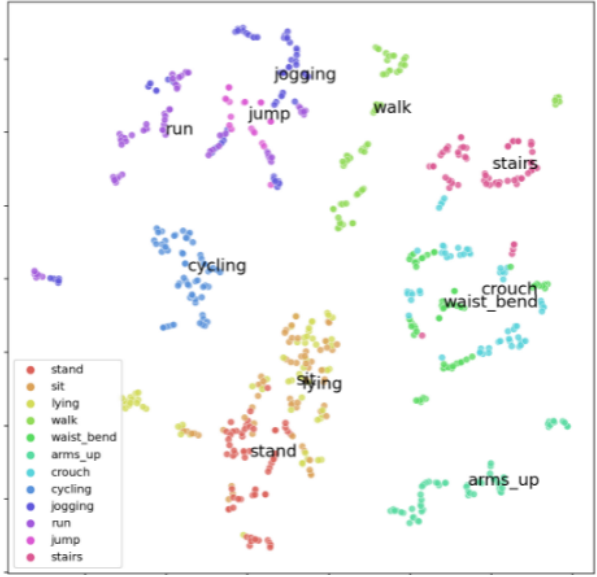

Daily Sports

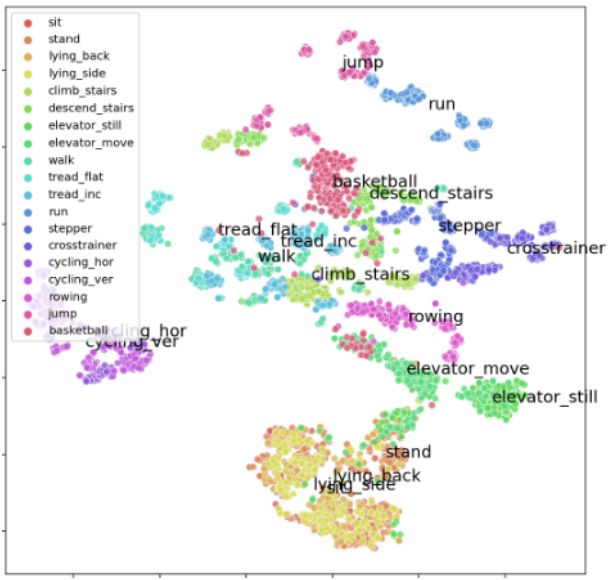

SupFig. 1. The t-SNE plots of activity representations of the 4 benchmark datasets.

2.

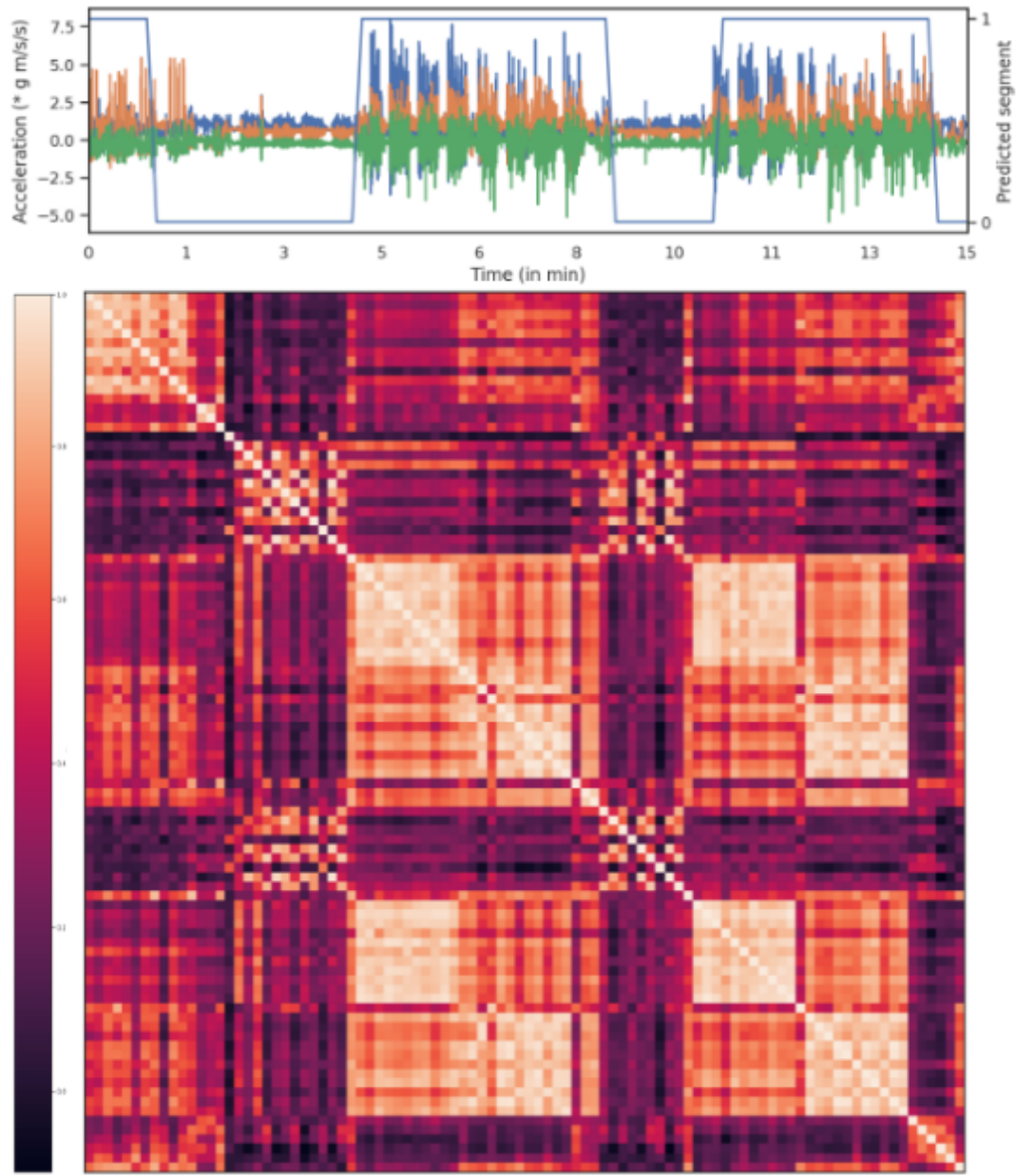

SupFig. 2 An example of salient activity segmentation labeled exercise. The first subfigure shows the IMU from the wrist-worn accelerometer scaled to units of g. Overlaid we show the salient activity segments predicted by our algorithm (value of 1 indicates a salient activity, 0 indicates indistinct motion). The second subfigures show the similarity matrix of the embeddings of 10 second windows of the same period. Element *ij* of the matrix is the cosine similarity between window i and window j, values closer to 1 indicate higher similarity than values closer to -1.

3.

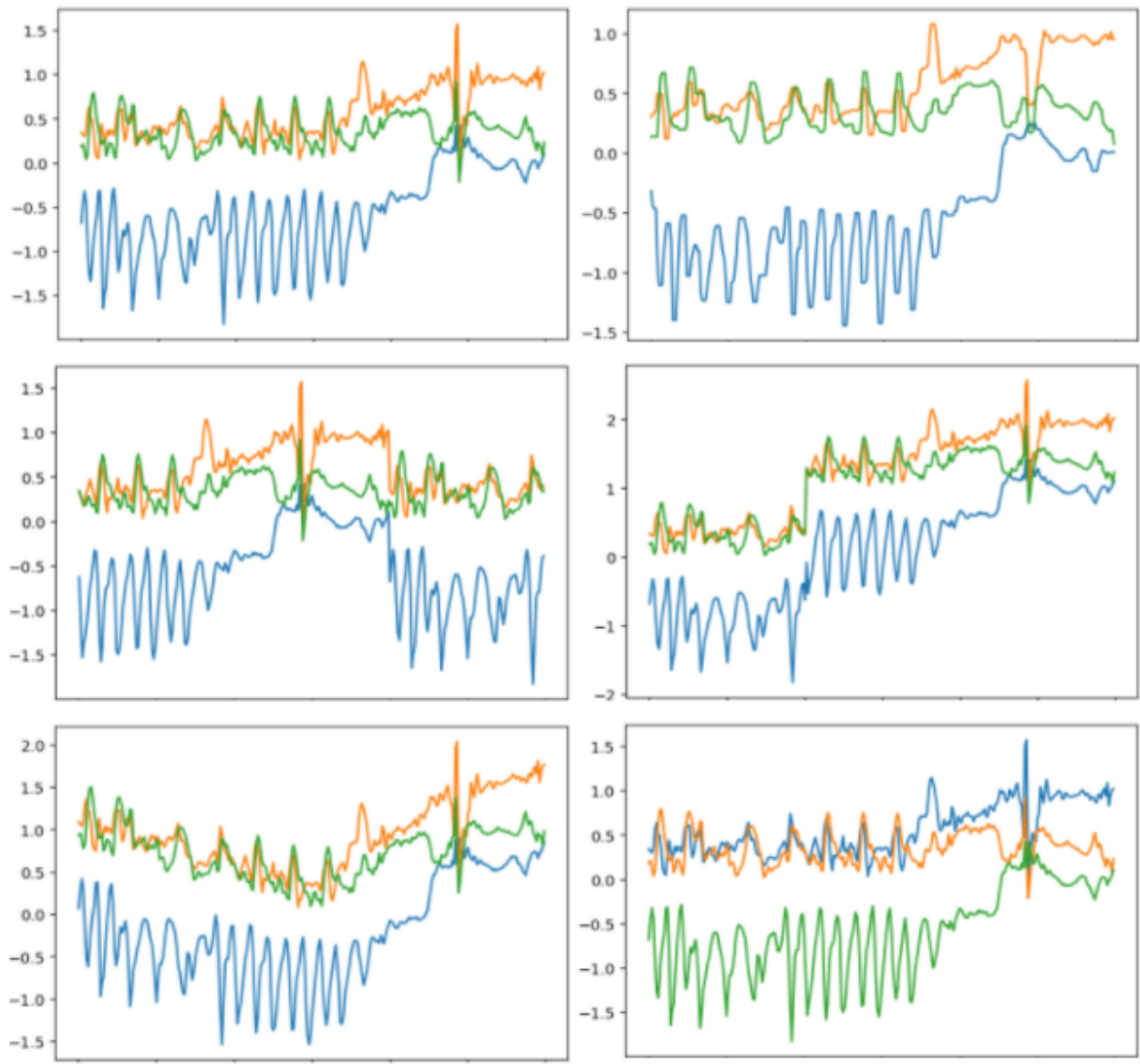

SupFig.3 Augmentations. a) is the original window and the rest are its augmented versions. b) is the smoothed window, c) has been translated in time, d) has a discontinuous jump in the baseline, e) has a gradual baseline wander, and f) is rotated in 3-axis.

# Acknowledgements

We thank Aren Jansen for invaluable guidance and feedback in designing the concept and implementation of self-supervised machine learning models. We thank the UCI Machine learning dataset repository for hosting the public benchmark datasets used in this effort. We thank all the participants in the Project Baseline Health Study for their participation.

Verily Life Sciences, LLC, funded the work and the research studies. All authors were full time employees of Verily Life Sciences during their contributions to this effort. No financial compensation was received outside of the contributors' regular monetary and stock compensation due to their employment at Verily Life Sciences.

# Author Information

## Affiliations

Verily Life Sciences, South San Francisco, CA, US

Niranjan Sridhar, Lance Myers

## Corresponding author

Correspondence to Niranjan Sridhar at nirsd@verily.com.

## Author contributions

N.S designed the technical concept, built the data infrastructure, the machine learning models, and the evaluation scripts and wrote the manuscript. L.M. served as scientific and technical advisor and supervised the project. All authors revised and approved the manuscript.

## Competing Interests

We declare no competing conflicts of interest for the authors.